\documentclass[conference]{IEEEtran}
\usepackage{amsmath,amssymb,amsfonts}
\usepackage{algorithm}
\usepackage{algpseudocode}
\usepackage{graphicx}
\usepackage{textcomp}
\usepackage{xcolor}
\usepackage{bbm}
\usepackage{hyperref}
\usepackage{comment}
\usepackage{makecell}

\newcommand{\re}[1]{{\color{black}#1}}

\usepackage{booktabs}
\usepackage{multirow}
\usepackage{booktabs}
\usepackage{soul}

\begin{document}

\title{VisionTraj: A Noise-Robust Trajectory Recovery Framework based on Large-scale Camera Network\\
\thanks{Identify applicable funding agency here. If none, delete this.}
}

\author{\IEEEauthorblockN{1\textsuperscript{st} Zhishuai Li}
\IEEEauthorblockA{\textit{SenseTime Research} \\
Shanghai, China \\
lizhishuai@sensetime.com}
\and
\IEEEauthorblockN{2\textsuperscript{nd} Ziyue Li}
\IEEEauthorblockA{\textit{Dept. of Information Systems} \\
\textit{University of Cologne}\\
Cologne, Germany \\
zlibn@wiso.uni-koeln.de}
\and
\IEEEauthorblockN{3\textsuperscript{rd} Xiaoru Hu}
\IEEEauthorblockA{\textit{SenseTime Research} \\
Shanghai, China \\
huxiaoru@sensetime.com}
\and
\IEEEauthorblockN{4\textsuperscript{th} Guoqing Du}
\IEEEauthorblockA{\textit{SenseTime Research} \\
Shanghai, China \\
duguoqing@sensetime.com}
\and
\IEEEauthorblockN{5\textsuperscript{th} Yunhao Nie}
\IEEEauthorblockA{\textit{China Telecom} \\
\textit{Cloud Technology Co., Ltd}\\
Beijing, China\\
nieyh@chinatelecom.cn
}
\and
\IEEEauthorblockN{6\textsuperscript{th} Feng Zhu}
\IEEEauthorblockA{\textit{SenseTime Research} \\
Shanghai, China \\
zhufeng@sensetime.com}
\and
\IEEEauthorblockN{7\textsuperscript{th} Lei Bai}
\IEEEauthorblockA{\textit{Shanghai AI Lab} \\
Shanghai, China \\
baisanshi@gmail.com}
\and
\IEEEauthorblockN{8\textsuperscript{th} Rui Zhao}
\IEEEauthorblockA{\textit{SenseTime Research} \\
and \textit{Qing Yuan Research Institute} \\
Shanghai, China \\
zhaorui@sensetime.com}
}

\maketitle

\begin{abstract}
Trajectory recovery based on the snapshots from the city-wide multi-camera network facilitates urban mobility sensing and driveway optimization.
The state-of-the-art solutions devoted to such a vision-based scheme typically incorporate predefined rules or unsupervised iterative feedback, struggling with \re{multi-fold challenges such as lack of open-source datasets for training the whole pipeline, and the vulnerability to the noises from visual inputs}. In response to the dilemma, this paper proposes VisionTraj, the first learning-based model that reconstructs vehicle trajectories from snapshots recorded by road network cameras.
Coupled with it, we elaborate on two rational vision-trajectory datasets, which produce extensive trajectory data along with corresponding visual snapshots, enabling supervised vision-trajectory interplay extraction.
Following the data creation, based on the results from the off-the-shelf multi-modal vehicle clustering, we first re-formulate the trajectory recovery problem as a generative task and introduce the canonical Transformer as the autoregressive backbone. Then, to identify clustering noises (e.g., false positives) with the bound on the snapshots' spatiotemporal dependencies, a GCN-based soft-denoising module is conducted based on the fine- and coarse-grained Re-ID clusters. Additionally, we harness strong semantic information extracted from the tracklet to provide detailed insights into the vehicle's entry and exit actions during trajectory recovery. The denoising and tracklet components can also act as plug-and-play modules to boost baselines.
Experimental results on the two hand-crafted datasets show that the proposed VisionTraj achieves a maximum +11.5\% improvement against the sub-best model. Furthermore, we explore the potential downstream applications, and our model still outperforms its peers. 
To facilitate the broader community research, we also release two separate datasets to mitigate the scarcity of related datasets. The code and data are available here \url{https://github.com/bonaldli/VisionTraj}.
\end{abstract}


\section{Introduction}

\begin{figure*}[t]
    \centering
    \includegraphics[width=\textwidth]{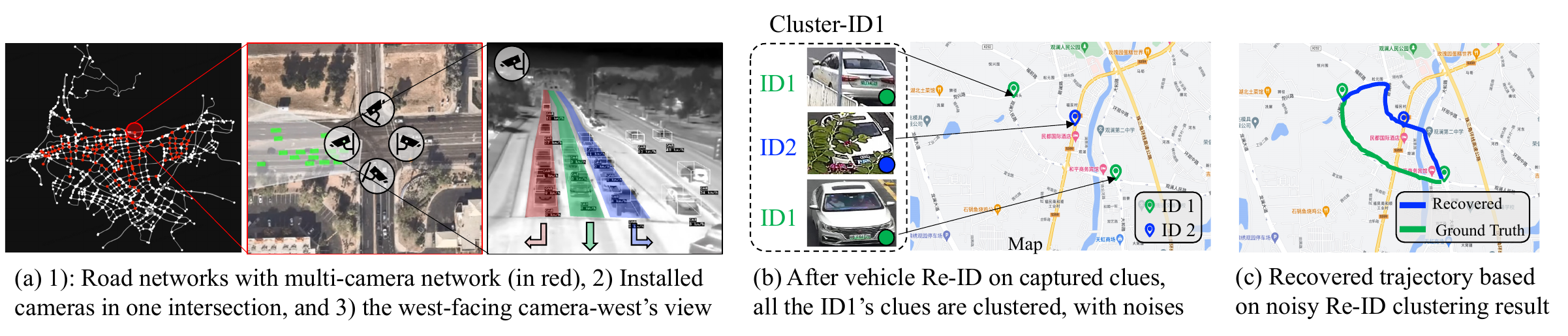}
    \caption{Introduction of multi-camera network and the sensitivity of trajectory recovery to Re-ID noises}
    \label{fig:intro}
\end{figure*}

Modern vision technology has revolutionized various industries. In the past few years, in the domain of intelligent transportation systems (ITS), several pioneering cities have started to install multi-camera networks to perceive, control, and optimize urban traffic. 
Usually, as shown in Fig. \ref{fig:intro}(a), four to six cameras are installed in each intersection facing each direction. Those cameras function as sensors to perceive the real-time traffic states, i.e., the immediate view of how many vehicles in each lane, and support various downstream traffic management tasks, such as calculating traffic indexes and traffic signal control. Camera-based sensing has many advantages  \cite{gordon2005traffic, fiore2019interdisciplinary}, and only a few of many are mentioned here: compared with traditional sensors such as loop detectors, the camera is with non-intrusive installment and cost-effective; compared with GPS, the camera can provide highly granular and precise data on a per-lane basis. 

To leverage the vision inputs from the camera network, a fundamental step is \textit{trajectory recovery}  \cite{wang_deep_2021,ren2021mtrajrec,chen2023rntrajrec}, \re{which is to recover a complete trajectory based on a sequence of discrete and rather sparse locations: those location records are obtained based on snapshots captured by the multi-camera networks}. Since cameras are usually only installed in a few intersections due to the hardware cost, \re{and in some extreme cases, the camera network only covers 30\% - 40\% of the intersections of a region.} Thus, how to link individual vehicles across multiple bypass intersections and recover their whole trajectory is required to enable further applications, such as general mobility pattern understanding \cite{9547354, li2021urban,ziyue2021tensor,9737052,li2022individualized,li2023tensor}, traffic prediction \cite{w:19, f:20, li2020long, chen2023adaptive,wang2023correlated,lin2023dynamic,jiang2023unified}, next location prediction \cite{l:21, zhang2022trajectory}, route planning \cite{wu2020learning},  travel time estimation \cite{f:20, yang2021unsupervised}, and more in-depth traffic index, including calculating travel time, average speed, and delay time.

Despite the advantages of cost-effectiveness and high-potential data quality, camera networks yet lay several algorithmic challenges to trajectory recovery.

First of all and most primarily, \textbf{there is no such complete public data available to support model training}. \re{To support the training of the whole pipeline with input as visual snapshots and output as the trajectories, the dataset should contain both the snapshots and the trajectories.} The main hindrance is that only a few developed cities have such a camera network infrastructure, and the data itself is also highly confidential. Our previous work from the colleagues, MMVC  \cite{lin2021vehicle} and CamTraj  \cite{yu2022spatio}, are the first team that published the representations of the original vision data with 5 million vehicle snapshots captured in Shenzhen, China. However, this data only offers the visual input and does not offer trajectory or the ground truth. \re{TrajData \cite{yu2023city} instead provided only the citywide trajectories obtained by the camera data in CamTraj  \cite{yu2022spatio}.} We keep improving the data further and release two high-quality datasets: (1) \textit{\textbf{Sewed-ViTraj Data}}: We sewed real vehicles' captured snapshots with simulated trajectories in a real city map. (2) \textit{\textbf{Simulated-ViTraj Data}}: we fully simulated a city (Longhua district in Shenzhen) with 157 intersections, 17 types of car models, 4875 cars, and in total 4875 trajectories. This could be the highest-quality data we could think of, given the protection and respect for data privacy. \re{The related datasets are compared in Table \ref{tab:data_summary}.}

Secondly and also most technically challenging, \textbf{the visual captures of vehicles inherently contain certain amounts of noise, which can dramatically affect the trajectory recovery quality}. We will first introduce the common pipeline in the industry to process the captured visual data, followed by the noise and how it will affect trajectory recovery significantly. As a common practice, the pipeline first preprocesses the visual features of the vehicles' appearances (using ResNet 50  \cite{he2016deep} or other vision deep models) and recognizes the car license plate (using Optical Character Recognition (OCR) models), then based on the visual features and spatiotemporal features, and it will put the snapshots into a same cluster if it thinks they are from a same car (i.e., one car is one cluster): this is also known in vehicle Re-ID \cite{liu2016large,khan2019survey}. After the Re-ID, as shown in Fig. \ref{fig:intro}(b), each snapshot is assigned a car ID, and the trajectory recovery model simply uses the same ID's locations to recover that vehicle's full trajectory. The challenges come from the upstreams of recovery: in real urban traffic, views of car plates can be blocked by obstacles, many cars could be in the same color or model, and weather conditions can also be bad: all these may result in the indistinguishable visual features of vehicles. As a result, false positives and false negatives in the Re-ID clustering will happen. Fig. \ref{fig:intro}(b) shows a false positive case of assigning vehicle ID-2 to the cluster of ID-1, because the ID-2's car plate is blocked by trees and ID-2 is in the same color and car model as ID-1. These are the inherent noises introduced from vision data. As a result, the trajectory recovery will be significantly affected, as demonstrated in Fig. \ref{fig:intro}(c). Commonly, false positives can render totally de-toured recovery results, and false negatives will cause inadequate samples to recover.

To the best of our knowledge, there are only a few works, i.e., MMVC  \cite{lin2021vehicle}, CamTraj  \cite{yu2022spatio}, TrajData  \cite{yu2023city}, and VeTraj  \cite{tong2021large}, that are tackling the same vision-based trajectory recovery problem. However, their solutions are still rather sensitive to the Re-ID noise because insufficient or even no attention was paid to the widely distributed and inherent noise. Moreover, from the model's perspective, these methods follow human-crafted or curated rule sets and identify noises with man-made assumptions, which are equipped with several rounds of unsupervised feedback \cite{lin2021vehicle,tong2021large} or prior knowledge such as historical trajectories \cite{yu2022spatio,yu2023city}. Thus, they usually lead to complexity with excessive rules and lag in the flexibility of learning or adapting. 

\begin{figure*}[t]
    \centering
    \includegraphics[width=1.0\textwidth]{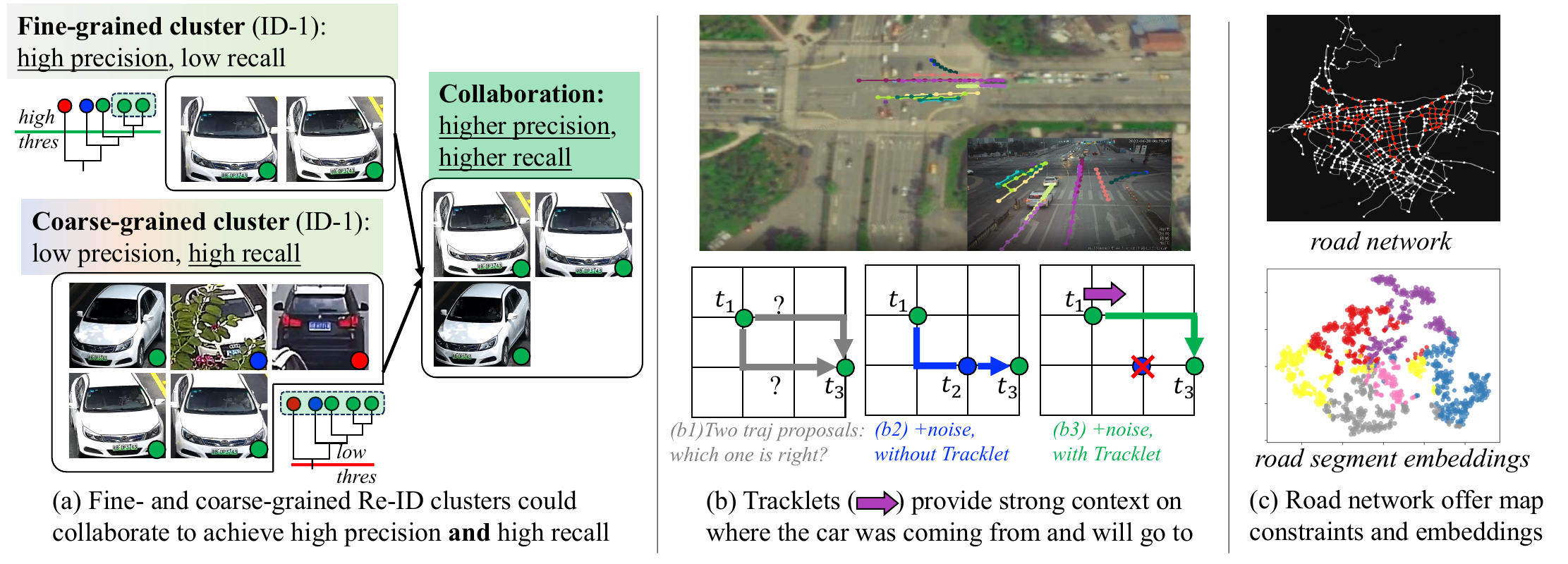}
    \caption{The key components introduced to tackle the Re-ID noise}
    \label{fig:key-components}
\end{figure*}

In this paper, we elaborate an unprecedented learning-based trajectory recovery model that enhances itself from data patterns without predefined rules or prior knowledge, and propose two innovative and effective components to tackle the challenge of Re-ID noise, as shown in Fig. \ref{fig:key-components}: (1) \textbf{Cooperation between fine- and coarse-grained Re-ID clusters}: in the Re-ID clustering process, we set two levels of similarity thresholds to decide clusters: shown in Fig. \ref{fig:key-components}(a), a higher similarity threshold gets a smaller cluster, with high precision but low recall (named as fine-grained); in parallel, a lower similarity threshold creates a bigger cluster, with high recall but lower precision (named as coarse-grained). We formulate these two clusters as two graphs and use the Graph Convolution Network (GCN) to encode the two graphs to learn a denoise score. (2) We propose to utilize the \textit{\textbf{Tracklets}} captured in each camera's local view, defined in Def. 3 in detail: as shown in  Fig. \ref{fig:key-components}(b), in case-(b1), there could be multiple indistinguishable trajectory proposals between two snapshots, and as shown in case-(b2), the existence of noise (the blue dot) will immediately mislead the recovery model to the blue trajectory. The tracklet (purple arrow) in case-(b3) instead provides strong contextual information on where this vehicle was coming from and where it goes to. This helps the model to identify the correct green trajectory and identify the blue dot as noise. In practice, we use the tracklet as a way of data augmentation, to generate the preceding and succeeding locations for a vehicle based on current intersection. 

From the perspective of trajectory recovery, we re-formulate the recovery as a generation task, i.e., the clustered records are regarded as prior information, and the generation model can yield a fine trajectory conditioned on it. The aforementioned information, denoising score from fine- and coarse-grained clusters, the augmented data based on tracklets, and road segment embedding will also be input to the trajectory generation model, using Transformer as the backbone.

The contributions are summarized as follows:

\begin{itemize}
    \item We propose a two-stage trajectory recovery framework composed of noise-robust vision clustering and learning-based trajectory generation. The proposed denoising module and tracklet augmentation are two universal plug-and-play modules, which can boost various baselines' performance. 
    \item A novel fine- and coarse-grain soft-denoising module is designed for false positive records identification from clusters. Overarching the canonical vision-based denoising, the module further exploits the spatiotemporal correlations of captured snapshots.
    \item The tracklet augmentation is elaborated with a dedicated bearing rate-based approach, which provides semantic details on the local behaviors of passing through vehicles under the cameras' field of view and thus contributes to the recall of recovered trajectories.
    \item To the best of our knowledge, we are the first to release a full dataset of the captured vision snapshots, along with the corresponding trajectories.
    It provides comprehensive information, including road networks, snapshots, camera locations,  tracklets under each camera, and vehicle trajectories.
    Moreover, we also exhibit a feasible pipeline for the construction of simulated data under the co-simulation by SUMO and CARLA. 
\end{itemize}

\section{Realted Work}
\subsection{Vision-based trajectory data}
Unlike mobility data such as GPS trajectory, the acquisition of vision-based vehicle trajectory is non-trivial, as it involves multiple sources from both vehicles and road networks: vehicles' travel paths should be tracked using devices such as GPS, while the corresponding snapshots need to be pinpointed from the camera-capture database when the vehicles pass through the intersections. 

\textbf{Real-world data}: Lin \textit{et al.}  \cite{lin2021vehicle} released an anonymized vision-based dataset, which includes multi-modality car features from 5 million captures. 
This is the only publicly available dataset known to us that contains both trajectory and capture information. Building upon this, in  \cite{yu2023city}, additional trajectories of the same region were introduced. However, these data lack corresponding capture information and are primarily used for speed pattern extraction of road segments. 
Due to the limited amounts of annotated trajectories (approximately 181), this real data cannot adequately propel learning-based trajectory recovery.

\renewcommand\arraystretch{1.5}
\begin{table*}[t]
\centering
\caption{The comparison of all available related datasets}
\resizebox{1.99\columnwidth}{!}{
\begin{tabular}{cccccccc}
\toprule
{\textbf{Dataset}} & {\textbf{City}} & {\textbf{\# Records}} & {\textbf{\# Cameras}} & \textbf{\# Trajectories} & \textbf{\# Road nodes} & \textbf{Snapshot input included?} & \textbf{Trajectory g.t. included?} \\\hline\hline
MMVC \cite{lin2021vehicle} & Shenzhen   & $\sim$5million                           & 673                               & 181  & 769  &  $\sim \text{(only embeddings)}$ &  $\checkmark$         \\\hline
CamTraj \cite{yu2022spatio} & Shenzhen   & $\sim$4million                           & 441                               & 197  & 1263  &  $\times$ &  $\times$         \\\hline
TrajData \cite{yu2023city} & Shenzhen~/~Jinan   & $\sim$4million~/~19million                           & 441~/~1838                             & 423~/~$\times$  & 1263~/~N/A  &  $\times$ &  \checkmark/$\times$         \\\hline
Sewed-ViTraj (Ours)        & Shenzhen   & $\sim$5million                           & 673                               & 25000  & 769  &  $\sim \text{(only embeddings)}$ &  \checkmark         \\\hline
Simulated-ViTraj  (Ours)    & Simulated Shenzhen   & 44307                             & 624                               & 4587   & 1839  &  \checkmark &  \checkmark   \\\bottomrule     
\end{tabular}}
\label{tab:data_summary}
\end{table*}

\textbf{Purely simulated data}:
Since labeling real captures is cumbersome, scholars have made attempts to automatically generate annotated data in simulation. One notable simulator is CARLA, known for its impressive realistic rendering capabilities. \cite{bullinger20183d} utilized it to generate 35 sequences of videos from five vehicles, which served as conducting in-camera trajectory recovery and scenarios reconstruction. However, this output incorporates only local motions under the field of view in cameras, rather than road network-level trajectories. TraCARLA  \cite{kumar2022vehicle} developed the simulated ReID dataset, which includes more than 50,000 captures taken from 85 cameras for over 700 different vehicle models. Based on the ReID output, the authors conducted the $A^*$ search algorithm with distance to heuristically reconstruct vehicle trajectories.
\cite{wang2022v2i} released the V2I-CARLA dataset, in which the authors built 5 traffic dual-camera scenes, and generated 1,500 captures for 153 vehicles with different colors and models. 
Created in Unity, \cite{yao2020simulating} introduced a large-scale synthetic dataset VehicleX, which contains 1,362 vehicle models with fully editable attributes. One can generate flexible amounts of captures with it for the ReID task.

To sum up, due to data privacy concerns and annotation barriers, obtaining large-scale real data to support downstream applications in trajectory is extremely challenging. Meanwhile, research on simulated data emphasizes capture quality/realism yet overlooks the capture-trajectory associated co-simulation, despite there are significant demands.
\subsection{Trajectory recovery based on multi-camera network}
The most relevant topic to trajectory recovery is multi-target multi-camera (MTMC) vehicle tracking \cite{liu2021city,yang2022box}, which has attracted sights of researchers and been presented in two consecutive CVPR competitions\footnote{https://www.aicitychallenge.org/}. However, in MTMC tracking, the cameras shall be deployed in consecutive adjacent intersections. In contrast, our task loosens restrictions that enable trajectory recovery even in lower camera coverage and wrong (noisy) ReID results.
When dealing with noise, several studies have introduced heuristic denoising methods, such as spatiotemporal constraints with multiple feedbacks  \cite{lin2021vehicle} and speed attributes of roads as probability/likelihood  \cite{yu2022spatio, yu2023city}.
VeTraj \cite{tong2021large} constructed a graph from the similarity of captures and utilized graph network properties for multimodal denoising.
The core implementations for trajectory recovery are relatively simple, usually by the $A^*$ search (Dijkstra) algorithm or Hidden Markov model (HMM) \cite{newson2009hidden}.

Yet another group involves reconstructing trajectories from plate-only data (i.e., license plate recognition results) rather than multi-modal vehicle attributes.  
 \cite{qi2021vehicle} proposes a TOPSIS-based decision algorithm for trajectory recovery, which finds the optimal trajectory by computing the weighted average of six metrics scores, such as path length and road hierarchy.
 \cite{wang2022general} takes the OD points as inputs to an LSTM model, generating a continuous path from the origin to the destination. However, the successful rate is low due to the poor routing capability of the sequence model.
 \cite{long2023vehicle} combines implicit temporal features within trajectories to investigate individual vehicle preferences. It employs knowledge representation learning to acquire embedded representations of trajectories and subsequently utilizes the embeddings to conduct trajectory link prediction, achieving the recovery task.

\subsection{General trajectory recovery based on mobility data}
Diverse sensing technologies have been applied to recognize vehicle mobility, such as GPS, unmanned aerial vehicles \cite{bock2020ind}, and RFID \cite{cheng2014virtual}. Among them, GPS is now the dominant provider for the perception of vehicles due to the increased demand for GPS-based navigation. 

To narrow down particular paths from low-sampled GPS trajectories, 
traditional methods exploit various heuristic algorithms, such as shortest path, query similar results from historical data, and HMM, and thus reduce the uncertainty between two consecutive sampled points in the trajectories.
In \cite{zheng2012reducing}, a history-based route inference system (HRIS) is proposed, which derives the travel pattern from historical movements and incorporates it into the route inference process.
\cite{ozdemir2018hybrid} proposes a hybrid HMM method to reconstruct the paths from low-sampled GPS, under the constraints of temporal discrepancy derived from legal speed limits or historical speed.

Recently, learning-based methods have been developed for trajectory recovery, such as DHTR \cite{wang2019deep}, MTrajRec \cite{ren2021mtrajrec}, JCRNT \cite{mao2022jointly}, MGCAT \cite{li2023critical}, and RNTrajRec \cite{chen2023rntrajrec}. 
They define the $\epsilon$-sampling interval trajectory, which refers to a time-ordered sequence of uniformly sampled points with specified time intervals. In this way, the sequence-to-sequence methods \re{such as RNN, LSTM, Transformer \cite{vaswani2017attention}, and BERT \cite{devlin2018bert,lewis2020bart}} can be introduced to conduct an end-to-end trajectory recovery, which employs an encoder to create the representation of a given trajectory and predicts a high-sampling trajectory by a dedicated decoder. \re{Moreover, self-supervised learning, such as contrastive learning \cite{chen2020simple}, has been further equipped to learn more general representations, e.g., JCLRNT \cite{mao2022jointly}.}

However, these methods are not suitable for our specific task: (1) visual and GPS inputs display distinct attributes. The former is more sparse, with noises and errors predominantly arising from visual recognition (e.g., failures in OCR or feature extraction). Conversely, the latter is primarily evident in device localization errors, essentially conforming to the Gaussian distribution. (2) When deployed in a new region, historical trajectories are unavailable for reference. (3) Ascertaining the number of missing trajectory points between two observed points for completion adds an additional layer of complexity.

\section{Our Approach}
\re{In this section, we will first give the preliminaries and important definitions in Sec \ref{sec: preli}; Then Sec \ref{sec: reid} will briefly introduce how snapshots from the same vehicle are clustered. Sec \ref{sec: model} gives the details of our model. Sec \ref{sec: train} introduces the training.}
\begin{figure*}[t]
    \centering
    \includegraphics[width=0.9\textwidth]{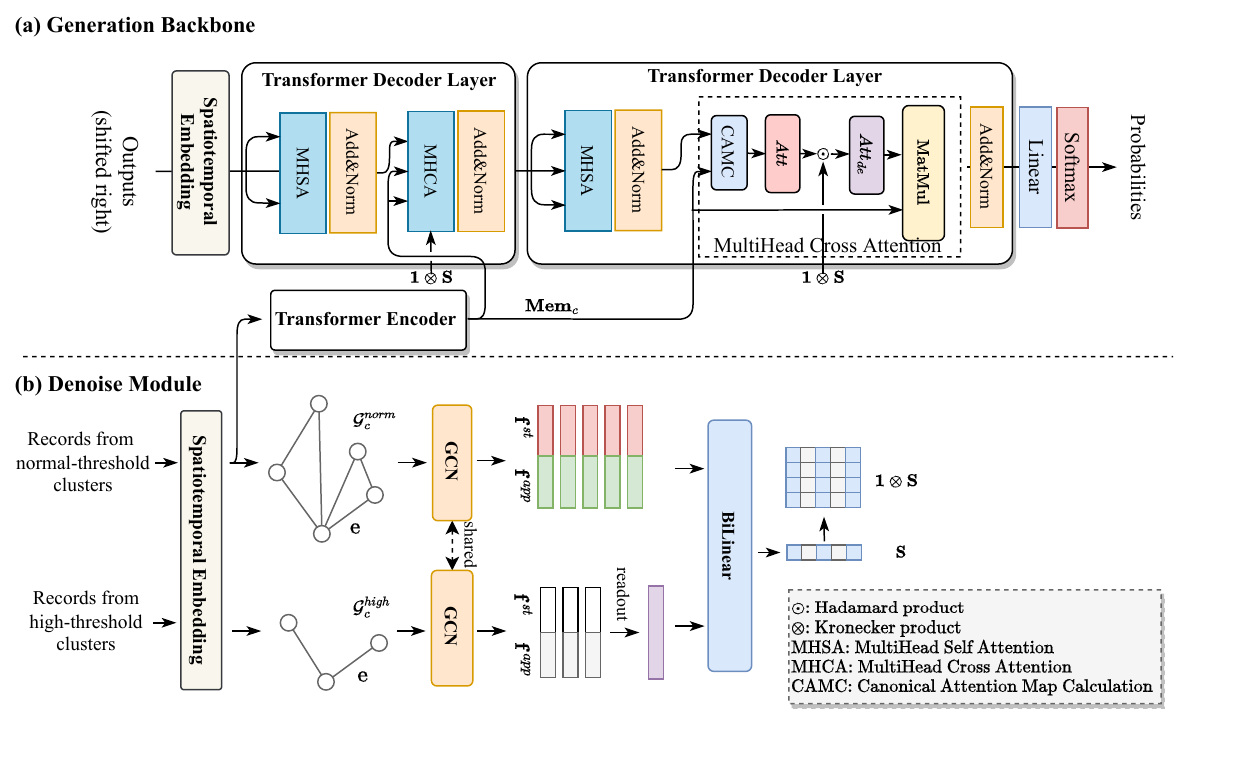}
    \vspace{-24pt}
    \caption{The overview of the proposed learning-based trajectory recovery model. 
    }
    \label{fig:overview}
\end{figure*}
\subsection{Preliminary}
\label{sec: preli}
\textbf{Definition 1}
Record $r_i$: The snapshot captured by the deployed camera when a vehicle passes through. It includes time stamp $t_i$, snapshot $s_i$, and camera id $cam_{i}$, that is, $r_i=[t_i, s_i, cam_{i}]$.

\textbf{Definition 2} Road node $n_i$: It is an intersection in the road network $\mathcal{G}_{net}=\{V, E\}$, where $V$ and $E$ are the set of vertices (intersections) and edges (road links), respectively. Each road node is represented by a unique ID allocated from 1 to $|V|$, that is, $n_i\in{1, 2, ...|V|}$. An intersection may deploy multiple cameras to face each entering approach (e.g., a 4-approach intersection usually has 4 cameras, but due to the cost, not all intersections are camera-equipped). According to the geo-locations of cameras and road nodes, we can relate the nearest road node for each camera. The record \re{defined in Def. 1} can thus be converted to $r_i=[t_i, s_i, n_{i}] $.

\textbf{Definition 3} Tracklet $tklet_i$: It refers to a segment of a vehicle trajectory in a specific area captured from the camera's local field view. When the snapshot record $r_i$ is triggered, the contextual video will also be archived. With tracking algorithms and \re{coordinate mapping} (\re{mapping the locations in the camera view to the coordinates in the real world}), the tracklet can be traced and extracted from camera video with the GPS coordinates. \re{Some tracklets are shown in Fig. \ref{fig:key-components}(b).}

\textbf{Definition 4} Trajectory $\tau_i$: A route that a vehicle follows. It can be represented by several types, such as latitude/longitude (GPS format). In this paper, instead, we use the consecutive and adjacent road nodes to depict the path, that is, $\tau_i=[n_1,n_2,...,n_M]$, where $M$ is the number of road nodes in $\tau_i$. The GPS-coordinate trajectories can also be translated into road node format by map-matching algorithms  \cite{meert2018hmm}. 

\textbf{Problem Formulation}
Given the snapshot database $\mathcal{D}=\{r_1, r_2, ..., r_{|\mathcal{D}|}\}$ that records $N$ vehicles $veh_i, i=\{1, 2, ...,N\}$, which pass through the road network $\mathcal{G}_{net}$ in a time range, the goal is to recover the trajectory $\hat{T}$ for each vehicle at the granularity of road nodes, which can minimize the distance between the recovered and ground truth trajectories. That is:
\begin{equation}
    Distance(T, {\arg\max}_{\hat{T}}~P(\hat{T}|\mathcal{D},\mathcal{G}_{net})), 
\end{equation}
where $T=\{\tau_1, \tau_2,... ,\tau_N\}$ contains the ground truth trajectory $\tau_i$ for each $veh_i$. $P(\cdot|\cdot)$ represents the likelihood of $\hat{T}=\{\hat{\tau}_1, \hat{\tau}_2,... ,\hat{\tau}_N\}$. 

\re{In the following sections, we will first briefly recap the process of vision-based clustering (vehicle Re-ID) in Section \ref{sec: reid}, then we introduce the key designs of fine- and coarse-grained clustering, tracklet, node embedding, and lastly, the trajectory recovery model.}

\subsection{Vision-based clustering}
\label{sec: reid}
The vision-based clustering aims to group similar records into the same cluster according to three modalities: appearance feature $\mathbf{f}^{app}$ (extracted by ReID model), carplate feature (extracted by carplate feature extraction model), and carplate registry number identification (extracted by carplate OCR model). The similarity of all records in the capture database is measured by analyzing these three factors through pairwise assessment.
In situations such as car plates getting blocked, the car plate fails to be captured; then the appearance will be the only useful measured criteria. The similarity scores between records and existing cluster centroids are pairwisely calculated. 
By setting a clustering threshold and inspecting the centroids of existing clusters, the record with similarity scores greater than the threshold is merged into the cluster, otherwise a new cluster with the record is created. As a result, all records in the same cluster are recognized as coming from the same vehicle. For more details, please refer to  \cite{lin2021vehicle}. 

\subsection{Trajectory generation framework}
\label{sec: model}
The overall framework is illustrated in Fig. \ref{fig:overview}, which consists of two components: (a) the GCN-based denoising module and (b) the trajectory recovery backbone. The denoising module operates based on the hierarchical clustering results. Subsequently, the denoising scores are passed into the autoregressive decoder phase within the backbone.
\subsubsection{Trajectory recovery backbone}
Given clustered clusters and records, we re-frame the problem and map it to a generative task. For the $c$-th cluster, we perform a sequence-to-sequence generation by Transformer  \cite{vaswani2017attention} conditioned on the road node $ n_i $ in records $r_i, i\in\{1, 2,...,|c|\}$ within it. 

Each road node can be handled as a word token, and the token embedding $\mathbf{x}_i$ is the spatiotemporal embedding of the record $r_i$, i.e., $\mathbf{x}_i=\mathbf{x}_i^{spat}+\mathbf{x}_i^{temp}$.
According to the captured timestamp $t_i$ for $r_i$, we adopt the time-encoding function to generate the temporal embedding $\mathbf{x}_i^{temp}\in\mathbb{R}^d$, for the $j$-th entry, there is
\begin{equation}
     (\mathbf{x}_i^{temp})_{j} = \cos \left(t_i \cdot \alpha^{-(j - 1)/\beta} \right),
\end{equation}
where $d$ is the embedding dimension and $\alpha=\beta=\sqrt{d}$.
Since the geo-location of record $r_i$ has been mapped nearest road node $n_i$, we introduce the Node2Vec \cite{grover2016node2vec} model to conduct unsupervised pretraining and encode the spatial embedding of the records so that the road network structure (e.g., connectivity and relative location) can be incorporated, that is
\begin{equation}
     \mathbf{x}_i^{spat} = Node2Vec(n_i).
\end{equation}

Based on the embedding lookup table, we input all the used node tokens into the Transformer encoder module and yield the memory features, \re{the direct output from the encoder, denoted as } 
$ \mathbf{Mem}_c \in\mathbb{R}^{|c|\times d}$ for $c$-th cluster. Next, we use a begin token  $\langle BOS \rangle$ to query $ \mathbf{Mem}_c $ and autoregressively decode road node sequences, until the end token $\langle EOS \rangle$ is encountered or the maximum length of the trajectory is reached. Each generation involves a multiclassification, and the number of classes is equal to 
$|\mathcal{G}_{net}|+1$, which means the number of road nodes in $\mathcal{G}_{net}$ plus the end token. For the $i$-th decoding step, the feedforward process can be represented by
\begin{equation}
\begin{aligned}
        &\mathbf{Mem}_c = {\rm TransformerEncoder}({\mathbf{x}_1, \mathbf{x}_2, ..., \mathbf{x}_{|c|}})\\
    &\mathbf{H}^{dec} = {\rm TransformerDecoder}(\mathbf{O}_i, \mathbf{Mem}_c)\\
    &Prob = {\rm Softmax}\left({\rm Linear}\left(\mathbf{H}^{dec}\right)\right)
\end{aligned},
\end{equation}
where $\mathbf{H}^{dec}$ means the hidden states from the decoder output. 
$\mathbf{O}_i$ is a list with the previous $i$ output tokens from the model with initial $\mathbf{O}_0=[\langle BOS \rangle]$.
$Prob$ represents the probabilities distribution of road nodes in $\mathcal{G}_{net}$.

\subsubsection{Denoising from the fine-coarse-grained clustering results}
In vision-based clustering, noises are inevitable. As the only distinguishable feature for vehicles of the same type (same model and same color) is the car plate, in cases where the car plate is fail-captured or unclear, the similarity calculated by the appearance-only feature between the vehicles will be high. Thus, they can be wrongly characterized into the same cluster.

Since the multi-modality vision information falls flat to handle such corner cases, we propose to couple with spatiotemporal dependencies of the captured records to denoise. The rationale lies in that all the records captured from a vehicle are bound to be on its trajectory, and the trajectory inherently holds spatiotemporal correlations, which can in turn verify whether the records are generated by the same vehicle.
\textbf{Besides the clustering results by a normal threshold, we re-cluster the snapped database with a higher threshold}. The results from it provide smaller recall yet larger precision (i.e., less noise), which can be regarded as the anchors to evaluate which records in the clusters with the normal threshold are noisy.

Based on the fine-coarse-grained clustering results, a GCN-based denoise module is designed, where each cluster from different thresholds is treated as a separate graph $\mathcal{G}_c=\{\mathcal{V}_c, \mathcal{E}_c\}, c\in \{1, 2,...,C\}$ ($\mathcal{G}^{norm}_c$ and $\mathcal{G}^{high}_c$ represent the clustering results with normal and higher thresholds, respectively) and the $i$-th record in $\mathcal{G}_c$ is regarded as a node $v_i\in \mathcal{V}_c$. The edge $e_{ij}\in \mathcal{E}_c$ is the vision similarity between $i$-th and $j$-th records.
\textbf{The denoise can be viewed as a node binary classification task} \re{for the snapshots on the} $\mathcal{G}^{norm}_c$, \re{by comparing with the high-precision graph $\mathcal{G}^{high}_c$'s information, where the snapshots on $\mathcal{G}^{high}_c$ are all considered as the correctly-clustered labels.}
The module inputs are $\mathcal{G}^{norm}_c$ and the matched $\mathcal{G}^{high}_c$, and its outputs are the denoise scores of all the records in the $\mathcal{G}^{norm}_c$, which represent the confidence ranging from 0 to 1, indicating whether the record (node) is noise.

Concretely, we use a shared 2-layer GCN model to perform node information aggregation for graphs $\mathcal{G}^{norm}_c$ and $\mathcal{G}^{high}_c$, respectively. The output $\mathbf{f}^{st}_i=GCN(\mathbf{x}_i)$ is then concatenated to the appearance feature $\mathbf{f}^{app}_i$ of record $r_i$, that is, $\mathbf{f}_i=[\mathbf{f}^{st}_i, \mathbf{f}^{app}_i]$.
After that, $\mathcal{G}^{high}_c$ is read out as an anchor to classify each node in $\mathcal{G}^{norm}_c$ by calculating pairwise similarity with bilinear transformation. The similarity acts as the denoise score $S_i$, that is
\begin{equation}
    S_i = \sigma(BiLinear(\mathbf{f}_i, \operatorname{Readout}(\mathcal{G}^{high}_c))), 
\end{equation}
where $\sigma(\cdot)$ is $\operatorname{Sigmoid}$ activation and average operator is chosen as  $\operatorname{Readout}$ function.

\textbf{The denoise score} $\mathbf{S}=[S_1, S_2, ..., S_{|c|}]$ \textbf{works in a soft-weighting way during the Transformer decoding phase}, i.e., making the query to decoder pay more attention to high-score tokens (non-noise) while ignoring low-score tokens (noise) from the encoder output (i.e., memory $\mathbf{Mem}_{c}$). 
Compared to using the classification results for hard denoising \re{(e.g., using a threshold over the similarity)}, soft denoising is differentiable and can be trained in an end-to-end manner while alleviating the error propagation issue.
And the cross-attention output $\mathbf{H}$ in the decoder is modified as: 
\begin{equation}
\begin{aligned}
    \mathbf{H} &= Att_{de} \mathbf{Mem}_{c}\\
    Att_{de} &= Att \odot (\mathbf{1} \otimes \mathbf{S})
\end{aligned}
\end{equation}
where $Att\in \mathbb{R}^{L\times|c|}$ is the canonical attention map which is calculated by the product of embeddings from query and key. $\odot \text{ and } \otimes$ means Hadamard and Kronecker product, respectively. $\mathbf{1}\in \mathbb{R}^{L\times 1}$ is an all-one vector and $L$ is the maximum generation lengh of the decoder.
\subsubsection{Tracklet extracted from video}
The tracklet can furnish details about the vehicle's entry and exit actions at intersections, including the directions of entry and exit. It can thus provide \re{strong semantic information}, i.e., upstream and downstream road nodes about the vehicle's passage through. However, as the tracklet is extracted with GPS coordinates and incompatible with road node-based trajectory recovery, we further adopt a bearing-based scheme to find up/downstream nodes from the tracklet.

Bearing measures the angle between the line connecting two consecutive points and true north. The bearing rate for point $p_i$ is the absolute difference between two consecutive bearings,
\begin{equation}
\begin{aligned}
    B_i    = &\left|\arctan\left(\frac{D([lat_i, lon_{i}],[lat_{i+1}, lon_{i}])}{D([lat_{i+1}, lon_{i}], [lat_{i+1}, lon_{i+1}])}\right)- \right. \\
    &\left.\arctan\left(\frac{D([lat_{i+1}, lon_{i+1}],[lat_{i+2}, lon_{i+1}])}{D([lat_{i+2}, lon_{i+1}], [lat_{i+2}, lon_{i+2}])}\right)\right|
\end{aligned}
\label{eq:bearing}
\end{equation}
where $D(\cdot, \cdot)$ is geodesic distance function. $lat$ and $lon$ mean latitude and longitude, respectively. 

\begin{figure}
    \centering
    \includegraphics[width = 0.85 \columnwidth]{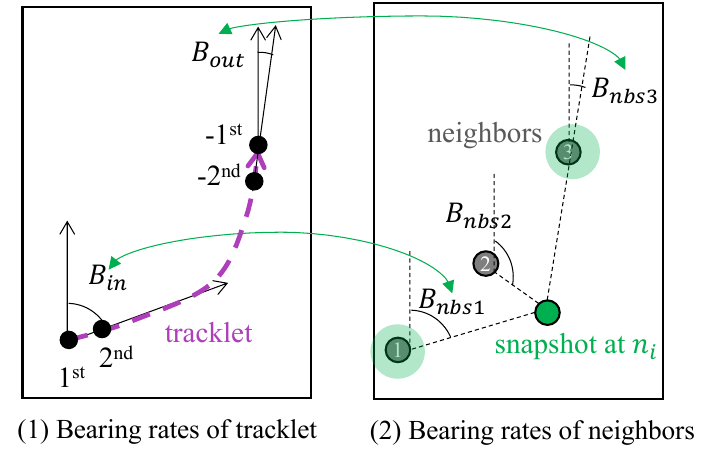}
    \caption{Search for the up/downstream road nodes by matching bearing rates}
    \label{fig:bearing-rate}
\end{figure}

The overall pipeline can be referred to in Alg. \ref{alg:bearing}. As illustrated in Fig. \ref{fig:bearing-rate}, from the extracted tracklet $tklet_i$ of record $r_i$, we first compute the entry bearing rate $B_{in}$ by the start and the second point, and the exit bearing rate $B_{out}$ by the end and the penultimate point. 
Next, we traverse all neighbors $\mathcal{N}(n_i)$ of road node $n_i$ and the bearing rate $B_{nbs}$ of the one which matches the $B_{in}$ or $B_{out}$ within a specified error~margin is identified as the up/downstream nodes for $n_i$. 
\begin{algorithm}[t]
\caption{Translate tracklet into up-down steam nodes}
    \begin{algorithmic}[1]
\State {$N_{tklet}=[n_i]$} \textcolor{teal}{{\textit{\small  
       \,\# node list defined for record $r_i$}}}
\State $\rm margin=20$ \textcolor{teal}{{\textit{\small  
       \,\# the threshold for evaluating the closeness between the points bearing and neighbors bearing}}}
\State $tklet_i$$=$ [$p_1$, $p_2$, ..., $p_{-2}$, $p_{-1}$] \textcolor{teal}{{\textit{\small  
       \,\# tracklet is defined by GPS-coordinated points}}}
    \State Calculate $B_{in}$ and $B_{out}$ by [$p_1$, $p_2$] and  [$p_{-2}$, $p_{-1}$], respectively. \textcolor{teal}{\textit{\small\# according to Eq. \eqref{eq:bearing}}}
    \For {$nbs \in \mathcal{N}(n_i)$} \textcolor{teal}{{\textit{\small  
       \,~\# search all the neighbors for $n_i$}}}
        \State Calculate $B_{nbs}$ by $nbs$ and $n_i$
        \If {$|B_{nbs} - B_{in}| \leq \rm margin$}
            \State $N_{tklet}$.insert(0, nbs) \textcolor{teal}{{\textit{\small  
       \,\# add upstream node $n^{up}_i$}}}
        \ElsIf{$|B_{nbs} - B_{out}| \leq \rm margin$}
            \State $N_{tklet}$.append(nbs) \textcolor{teal}{{\textit{\small  
       \,\# add downstream node $n^{down}_i$}}}
        \EndIf
    \EndFor\\
\Return $N_{tklet}=[n^{up}_i, n_i, n^{down}_i]$
\end{algorithmic}
\label{alg:bearing}
\end{algorithm}

After extracting up/downstream road nodes by tracklet $tklet_i$, we further augment the original node of the record $r_i$ by adding them into the node list, i.e., $[n_i^{up}, n_i, n_i^{down}]$. We feed this expanded node list directly into the Transformer model. Note that the scores from the denoising module also need to be extended accordingly, which is performed by repeating the denoise scores (e.g., {$repeat\_interleave(\mathbf{S}, 3)$}).
\subsection{Model Training}
\label{sec: train}
The generation backbone and the denoise module are co-trained during the training phase. Given that they are both classification tasks, we employ cross-entropy loss $CE$ for supervised learning and tune the parameters. For the $j$-th round generation, the generation loss is calculated by $\mathcal{L}_{gen}=CE(Prob,\tau_{i,j})$, where $\tau_{i,j}$ means the $j$-th token (road node) in ground truth trajectory $\tau_{i}$. For denoising, $\mathcal{L}_{de}=CE(S_i,S^*_i)$, where $S^*_i=0$ if $r_i$ in the normal cluster is a noise otherwise $S^*_i=1$. Finally, the total loss $\mathcal{L}$ for co-training is $\mathcal{L}=\mathcal{L}_{gen}+\lambda\mathcal{L}_{de}$ and we set $\lambda=1$ in this paper.
\section{Data Preperation} 
Due to the proposed trajectory recovery being learning-based, in this section, we provide a detailed explanation of how the training data is designed. As mentioned, we designed two datasets: Sewed-ViTraj data and Simulated-ViTraj data. 
Over the course, we consider how to construct reasonable trajectory datasets from two distinct scopes: (1) how to simulate noise distribution in the clustering results, and (2) how to produce ground-truth trajectories. 
Next, we will describe their fundamentals and rationales in detail. 
\subsection{Sewed-ViTraj Data}
We first recap the capture database from real camera networks from the published dataset in pioneering work  \cite{lin2021vehicle}, which records the representations of the original vision data with 5 million vehicle snapshots about 150 intersections in Shenzhen, China.
Then, we group them by the vision-based clustering algorithm and try to construct a simulated trajectory for each cluster. The noises and trajectories are specified by heuristic rules that are close to real scenarios.
Concretely, to accommodate the aforementioned scopes, we use two independent schemes to generate the trajectories separately.

\begin{figure}[t]
    \centering
    \includegraphics[width=\columnwidth]{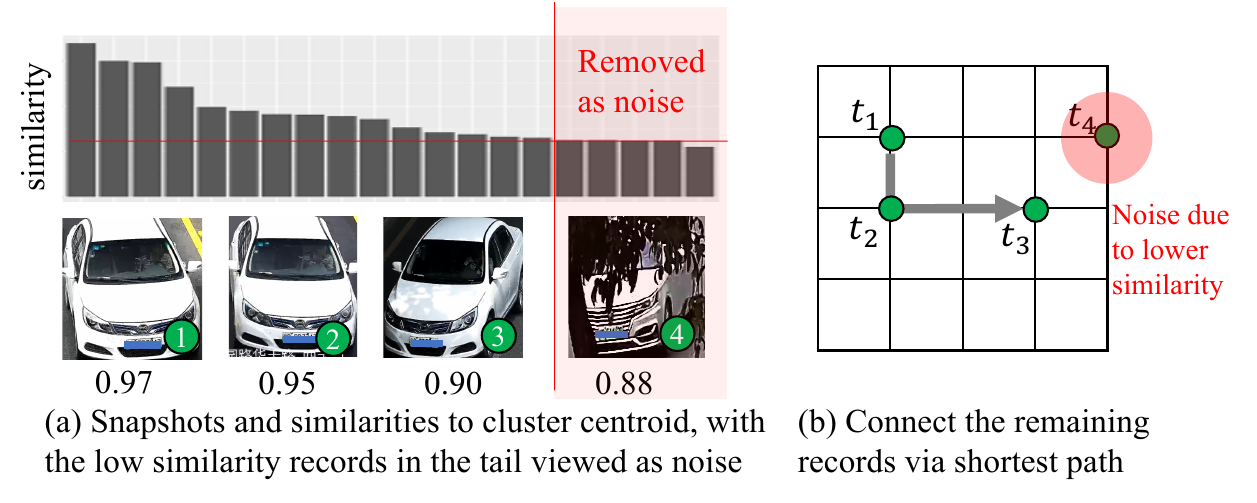}
    \caption{The illustration of \textbf{Scheme-1}, which first identifies the low-similarity records in the cluster as false positives and then concatenates the remaining records by the shortest path as the trajectory.}
    \label{fig:sheme1demo}
\end{figure}

\textbf{Scheme-1: shortest path trajectory across the true positive records in a cluster.} 
As Fig. \ref{fig:sheme1demo} illustrates, in this scheme, low-similarity captures in the clusters are marked as noise, and the true positives are then subsequently connected by shortest paths as simulated trajectories. The rationale behind this is that if noise is present, captures with low similarity within the cluster are more likely from other vehicles, even if the similarity has exceeded the clustering threshold. \re{We obtain 20,000 trajectories via the Shceme-1.}

\begin{figure}[t]
    \centering
    \includegraphics[width=\columnwidth]{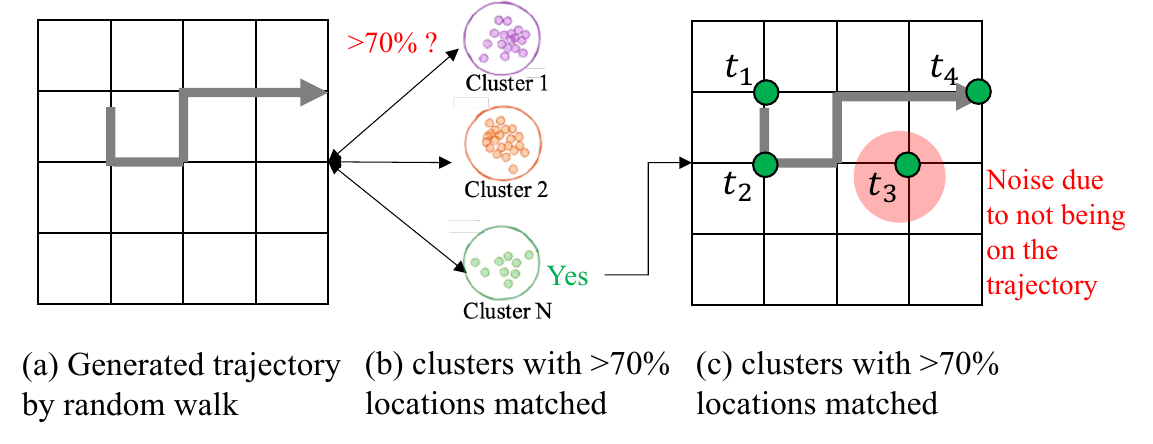}
    \caption{The illustration of \textbf{Scheme-2}, which first generates trajectory and then identifies the outside records as false positives in the cluster.}
    \label{fig:sheme2demo}
\end{figure}

\begin{figure}
    \centering
    \includegraphics[width=0.65\columnwidth]{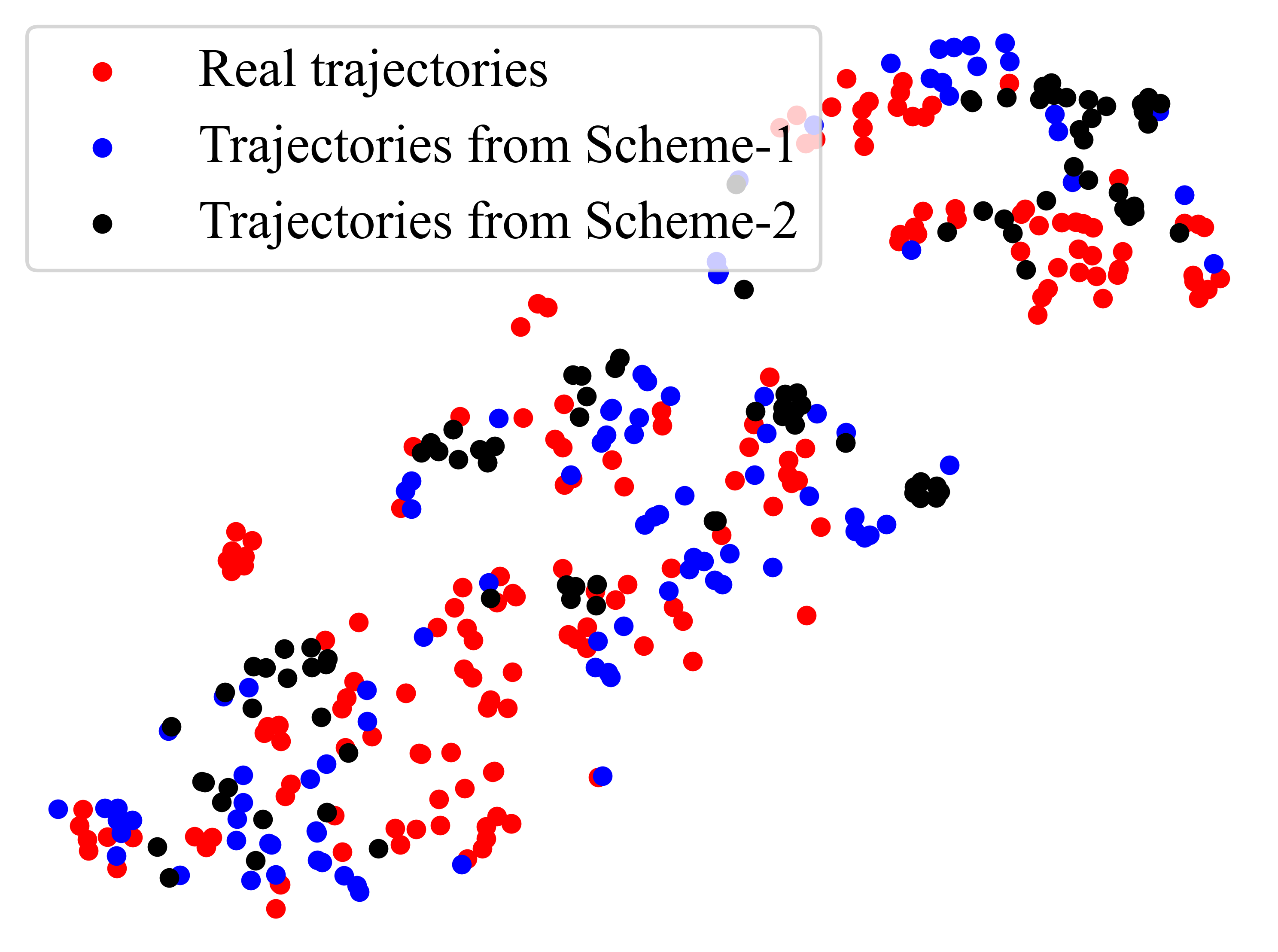}
    \caption{Visualization of two simulated trajectories from Scheme-1 and -2 v.s. the real trajectories: \re{the highly-overlapped distributions of the two schemes and the real trajectories justify the rationality of two schemes.}}
    \label{fig:emb_traj}
\end{figure}

\textbf{Scheme-2: random walk trajectory that passes through some samples in a cluster.}
In contrast to the former, this scheme first generates a trajectory by random walk and then looks up clusters with at least 70\%  of the locations of records therein matching the trajectory. In Fig. \ref{fig:sheme2demo}, the records with the locations on the trajectory are labeled as correct; otherwise, noise. This is motivated by the fact that whether a record is noise or not, it may be distributed randomly, and it doesn't have to be located at the tail part of the similarity distribution, e.g., when there is only an appearance similarity criteria for the records without carplate information, captures with high similarity may be noise. \re{We obtain 5,000 trajectories via the Shceme-2. The reason for a smaller amount is partially the computational cost when matching a random walk-generated trajectory to real snapshot clusters.}

By the two schemes, we get 25,000 trajectories based on the 5 million records. \re{As shown in Fig. \ref{fig:emb_traj}, we use t-SNE to visualize all the trajectories from Scheme-1, -2, and real ones, and we can see their distributions are quite overlapped, proving the rightness of our simulation.} Since the dataset is constructed by real images without raw videos, tracklet can not be extracted directly. Alternately, 
considering that the role of the tracklet is to suggest the up/downstream road nodes that a vehicle passes through, we find them from the ground truth trajectory and append them to the node list. In case a record does not attend to a trajectory, we randomly select its neighboring road nodes as the node list. 
\subsection{Simulated-ViTraj Data} 
We also exclusively generate city-wide vehicle records and trajectories through co-simulation by only SUMO\footnote{\url{https://sumo.dlr.de/docs/index.html}} and CARLA\footnote{\url{https://carla.org/}}. In this dataset, the noise in clusters is entirely caused by similarity errors when clustering rather than specified by rules like Sewed-ViTraj. The trajectories are automatically generated by the SUMO routing algorithm, taking into account the configured origins and destinations.
\begin{figure}[t]
    \centering
    \includegraphics[width=1\columnwidth]{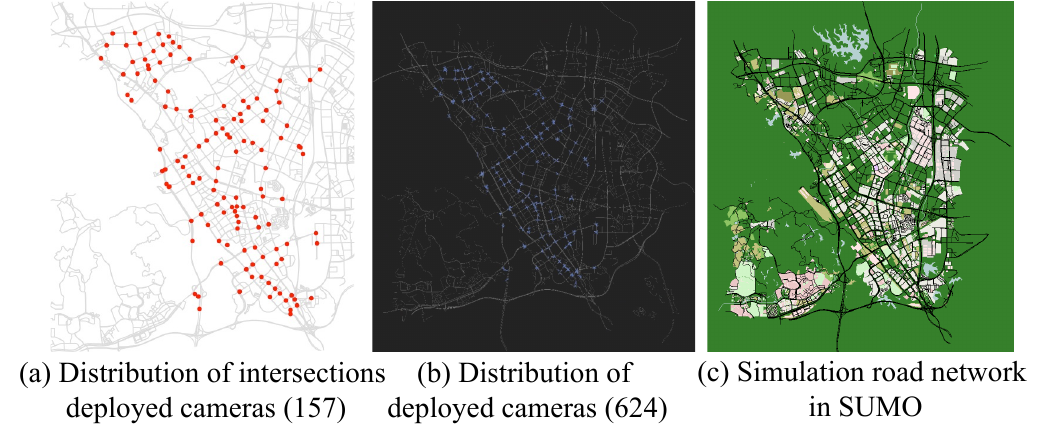}
    \caption{The distribution of intersections and cameras 
    }
    \label{fig:dist}
\end{figure}

\begin{figure}[t]
    \centering
    \includegraphics[width=0.85\columnwidth]{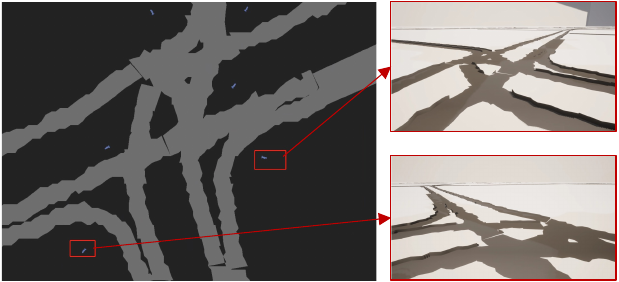}
    \caption{Two cameras and their fields of view in CARLA}
    \label{fig:cam_demo}
\end{figure}

\begin{figure}[t]
    \centering
    \includegraphics[width=0.90\columnwidth]{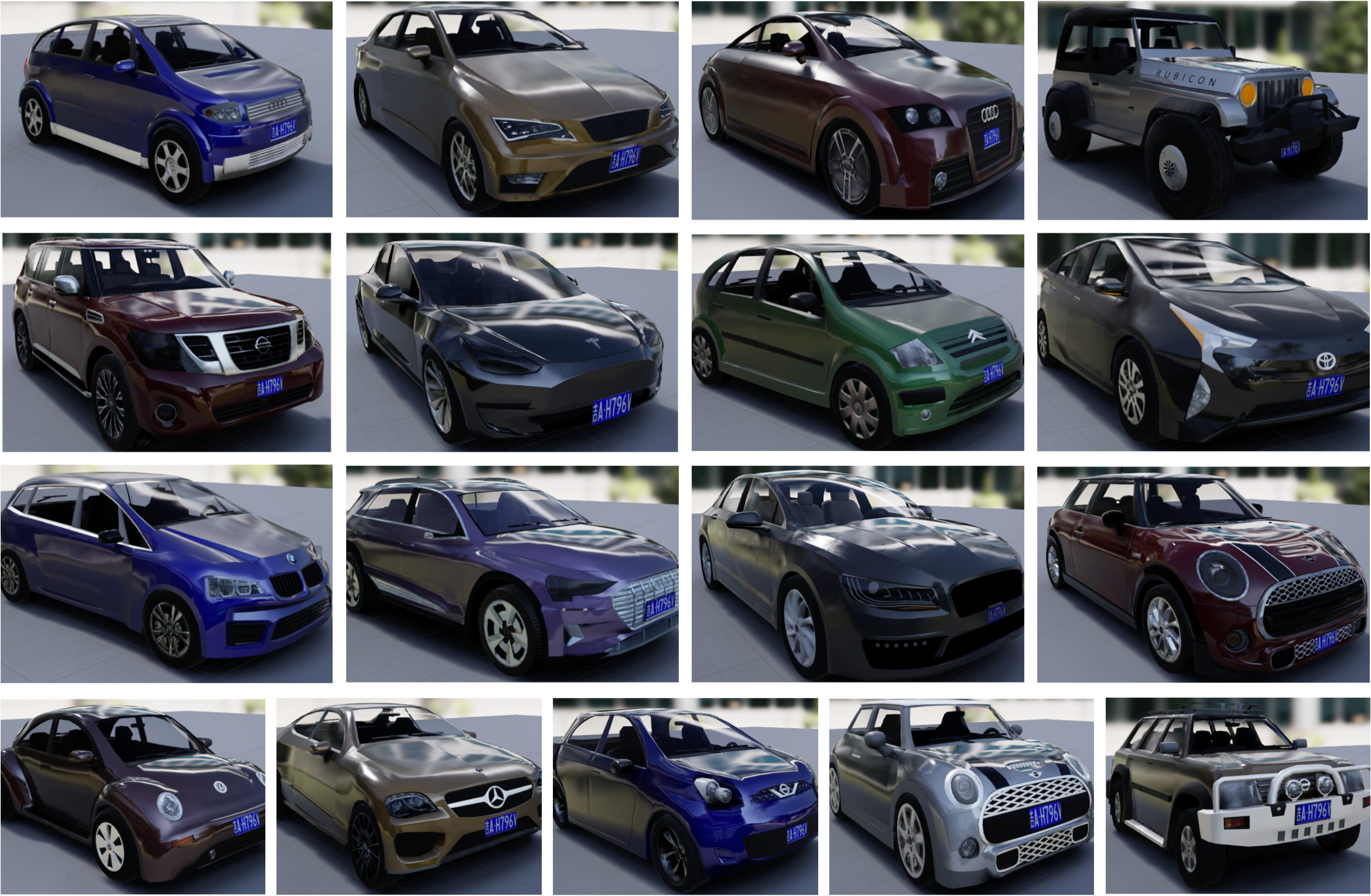}
    \caption{17 types of vehicles used for simulation}
    \label{fig:car_types}
\end{figure}
\begin{figure}[t]
    \centering
    \includegraphics[width=0.75\columnwidth]{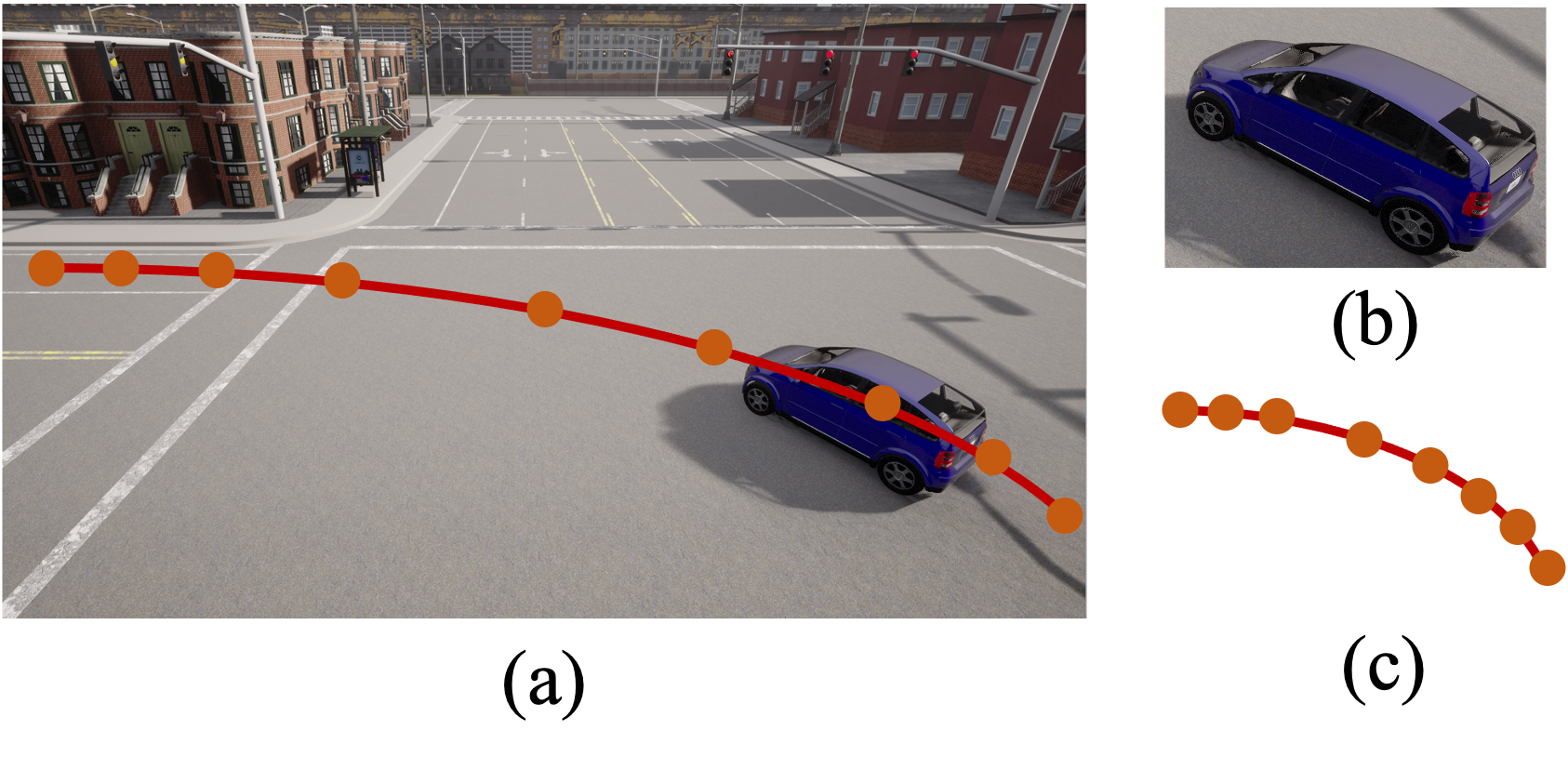}
    \caption{The captured image processing illustration. (a) The captured full-size image. (b) Cropped with a bounding box for feature extraction. (c) Extracted tracklet in this intersection.}
    \label{fig:cap_ill}
\end{figure}
\begin{figure}[t]
    \centering
    \includegraphics[width=0.95\columnwidth]{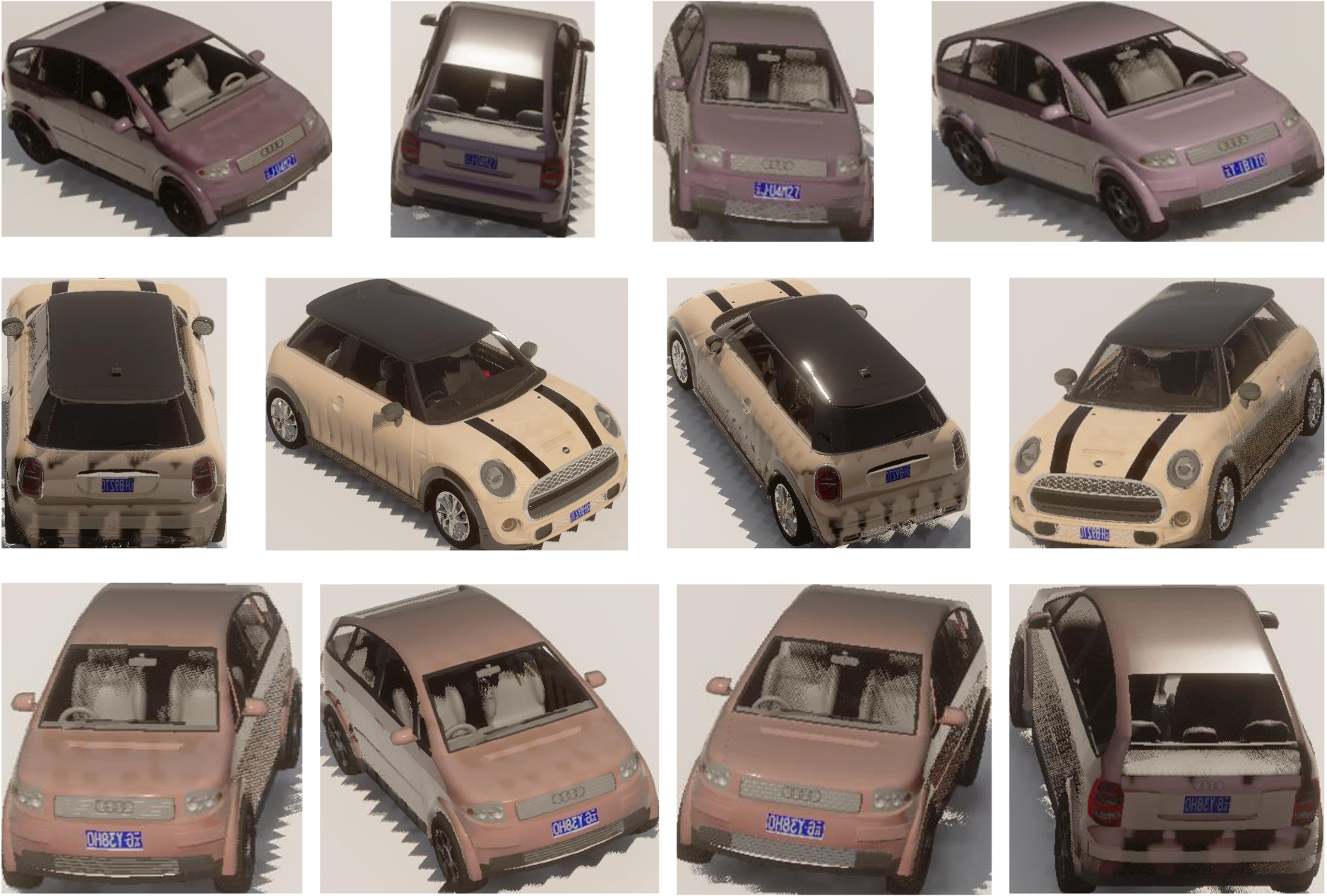}
    \caption{These snapshots showcase images captured from three vehicles, with each row depicting the images from the same vehicle.}
    \label{fig:sim_cap_img}
\end{figure}

\textbf{Simulation settings:} We manually deploy 624 cameras at 157 intersections in Longhua District, Shenzhen. Their distribution is shown in Fig. \ref{fig:dist}. Two cameras and their fields of view are also demonstrated in Fig. \ref{fig:cam_demo}. 
To ensure synchronization in map data during co-simulation, the road network is initially extracted from OpenStreetMap, resulting in a file with \texttt{.osm} format, which serves as the foundational map for SUMO. Subsequently, this identical map is then converted into OpenDRIVE format and imported into CARLA.
A total of 17 types of vehicles are deployed and their colors and license plate numbers are randomly and reasonably generated. Their brands and models are shown in Fig. \ref{fig:car_types}. 
The SUMO is used to configure vehicle origin-destination settings, generate trajectories, and simulate traffic signal control schemes. CARLA is employed to obtain vehicle snapshot data, tracklet, and other related information. 
As Fig. \ref{fig:cap_ill} illustrates, upon a vehicle's entry into the camera's field of view, a full-sized image is captured, and a cropped image is concurrently generated and stored based on the vehicle detection bounding box, creating a single vehicle record. Furthermore, the corresponding tracklet can also be recorded within the camera views.

\begin{figure}[t]
    \centering
    \includegraphics[width=0.65\columnwidth]{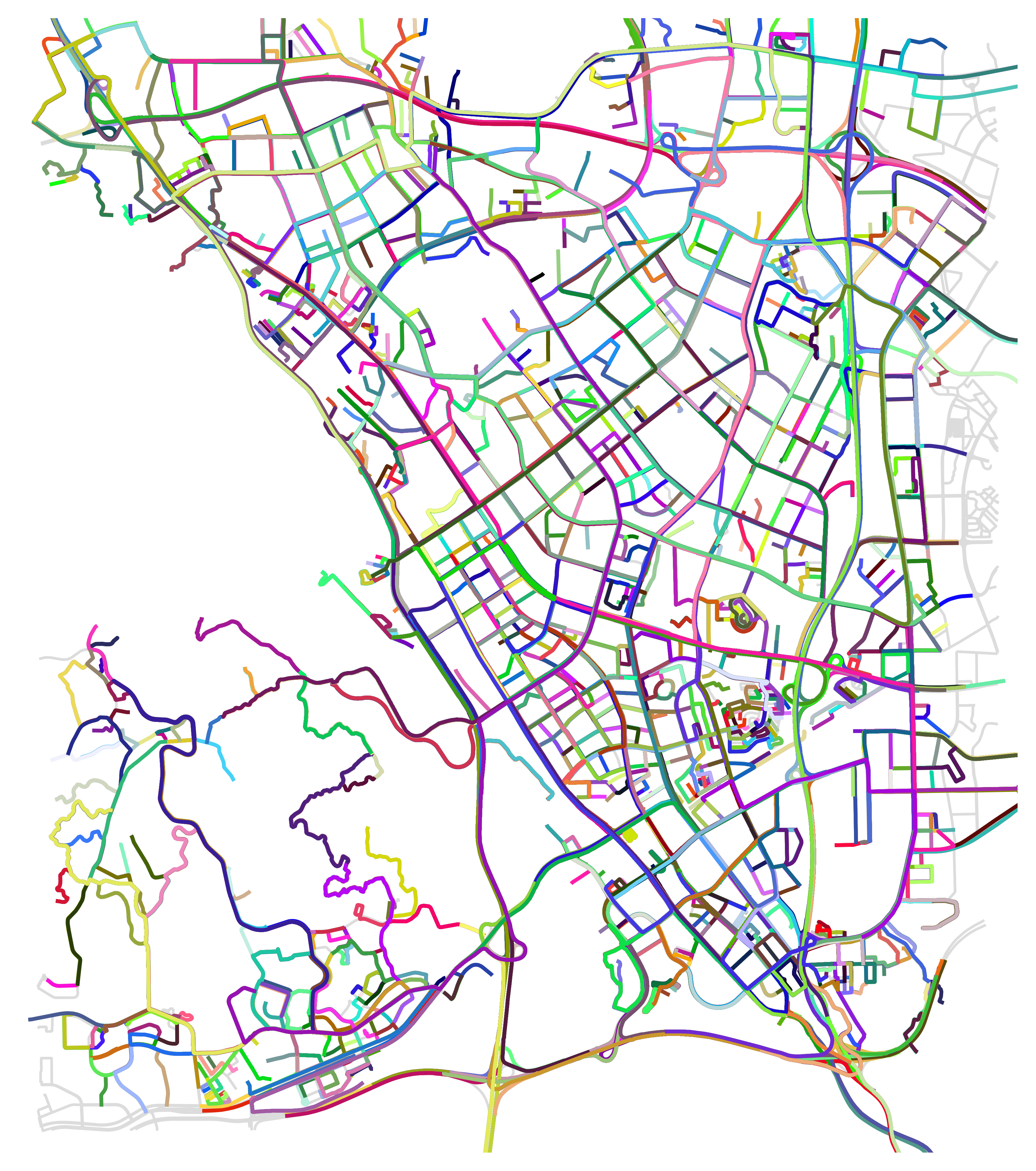}
    \caption{The visualization of the simulated road network (gray lines) and 4,875 trajectories, where each colored line represents the unique trajectory from a vehicle.}
    \label{fig:traj_plot}
\end{figure}
\textbf{Data attributes:} Under 3,600 seconds simulation time steps, we simulate a total of 4,875 vehicles and 4,875 trajectories under 60$km^2$ region, resulting in 44,307 capture images. We also visualize the samples of captured images from three vehicles in Fig. \ref{fig:sim_cap_img}, and all the trajectories in the road network are plotted in Fig. \ref{fig:traj_plot}. Subsequently, we processed the capture data by employing ResNet-50 for capture feature extraction, license plate feature extraction, and OCR license plate recognition.

\section{Experimental Analysis}

\subsection{Dataset Description}
In Table \ref{tab:datadesc}, we provide an overall summarization of the two constructed datasets.  
For the two datasets, we randomly select 512 and 50 samples (clusters) for testing, respectively, and the remaining for training. For Sewed-ViTraj, we choose 0.75 and 0.9 as the normal and high thresholds for clustering, respectively. For Simulated-ViTraj, 0.8 and 0.9 are opted as the thresholds.
Trajectory recovery is only conducted on clusters containing more than three records, as the clusters with fewer records are insufficient to support the task.
\renewcommand\arraystretch{1.5}
\begin{table}[h]
\centering
\caption{The basic statistics of two datasets}
\resizebox{\columnwidth}{!}{
\begin{tabular}{ccccc}
\toprule
{\textbf{Dataset}} & {\textbf{\# Records}} & {\textbf{\# Cameras}} & \textbf{\# Trajectories} & \textbf{\# Road nodes} \\\hline\hline
Sewed-ViTraj           & 4893350                           & 673                               & 25000  & 769            \\\hline
Simulated-ViTraj       & 44307                             & 624                               & 4587   & 1839     \\\bottomrule      
\end{tabular}}
\label{tab:datadesc}
\end{table}
\subsection{Baselines}
After the clustering process in \cite{lin2021vehicle}, the results are used for trajectory recovery. Based on the benchmark, we compare our method with several baselines:

(1) \textbf{SP} \cite{sniedovich2006dijkstra}: Shortest path.  It arranges all the captured images in the cluster in chronological order and connects their locations by the Dijkstra algorithm.

(2) \textbf{SP+tklet}: SP with tracklet. The road nodes extracted from the tracklet, both upstream and downstream, are added to the road node list, and the SP algorithm is then used to recover trajectories based on it.

(3) \textbf{HMM} \cite{newson2009hidden}: Hidden Markov model. It performs map matching based on the distance-based road network.

(4) \textbf{DHM}: Denoising module with hard margin. We customize our denoising module by a hard threshold of 0.5 to determine the confidence of records within the cluster. The record with $S < 0.5$ will be considered as noise; otherwise, it will be classified as true positive. The denoising module thus becomes a plug-and-play component that can be migrated to other powerful baselines. Then, the SP or HMM algorithm is performed only on the records after the hard denoising, \re{denoted as \textbf{SP+DHM$^\text{de}$} and \textbf{HMM+DHM$^\text{de}$} respectively.}

(5) \textbf{MMVC (w. ST$^\text{de}$)} \cite{lin2021vehicle}: Multi-modality vehicle clustering with their proposed denoising feedback. It first uses HMM to recover the trajectory based on the clustering results and then denoises some of the false positives by the trajectory-level spatial-temporal information.
\subsection{Evaluation Metrics}
The objective of the task is to reconstruct road node-represented trajectories from captured records. Consequently, we employ the Precision (P), Recall (R), and Intersection over union (I) on the road node level to evaluate the performance of our model and the baselines, all of which are the higher, the better. The three criteria compare two sets of road nodes from predicted trajectory $\hat{\tau}$ and ground truth ${\tau}$, which are formulated~as
\begin{equation}
\begin{split}
& \text{Precision} = \frac{1}{N_T}\sum_i^{N_T}\frac{|\hat{\tau_i}\cap\tau_i|}{|\hat{\tau}_i|},\\
& \text{Recall} = \frac{1}{N_T}\sum_i^{N_T}\frac{|\hat{\tau_i}\cap\tau_i|}{|\tau_i|},\\
& \text{IoU} = \frac{1}{N_T}\sum_i^{N_T}\frac{|\hat{\tau_i}\cap\tau_i|}{|\hat{\tau_i}\cup\tau_i|}
\end{split}
\end{equation}
where $N_T$ means the number of testing data. $\hat{\tau_i} \cap \tau_i$ represents the set of true positive road nodes in the $i$-th prediction. $|\cdot|$ indicates the set cardinality operator.

\begin{table*}[t]
\centering
\caption{Performance comparison. The best results are in \textbf{boldface} and the second-best ones are \underline{underlined}.}
\resizebox{\textwidth}{!}{
\begin{tabular}{c|c|cccccccc}
\toprule
        \textbf{Datasets}            &   \textbf{Metrics}   & SP    &   SP+tklet &SP+DHM$^\text{de}$ & SP +DHM$^\text{de}$+tklet&HMM   & HMM+DHM$^\text{de}$ & MMVC (w. ST$^\text{de}$)  & ours  \\\midrule
\multirow{3}{*}{\textbf{Sewed-ViTraj}} & \textbf{Precision} & 0.822 &    0.793  & 0.825      &0.799  &0.829 &\underline{0.831}       & -     & \textbf{0.903} \\\cline{2-10}
                    & \textbf{Recall}    & 0.883 &    \underline{0.912}   &0.867      & 0.898 &0.873 &0.860       & -     & \textbf{0.917} \\\cline{2-10}
                    & \textbf{IoU}       & 0.749 &    0.736   &\underline{0.749}      & 0.730 &0.749 &0.745       & -     & \textbf{0.836} \\\midrule
\multirow{3}{*}{\textbf{Simulated-ViTraj}} & \textbf{Precision} & 0.860 &    0.773  &\textbf{0.901}      & 0.800  &0.864 &0.885       & {0.865} & \underline{0.896} \\\cline{2-10}
                    & \textbf{Recall}    & 0.919 &    \textbf{0.933}   &0.891      & 0.913 &0.900 &0.864       & 0.909 & \underline{0.924} \\\cline{2-10}
                    & \textbf{IoU}       & 0.824 &    0.746   &\underline{0.838}      & 0.756 &0.811 & 0.804       &0.820 & \textbf{0.852}\\\bottomrule
\end{tabular}}
\label{tab:overall_perf}
\end{table*}
\subsection{Trajectory recovery results}
\textbf{Performance:} In Table \ref{tab:overall_perf}, we summarize the overall performance on trajectory recovery across the baselines, with the best in boldface and the second-best underlined. 

The results indicate that our proposed model surpasses the baselines, showcasing a significant improvement across both datasets. In comparison to MMVC (w. ST$^\text{de}$), our method demonstrates a notable enhancement ranging from 1.6\% to 3.9\% under the three criteria. Particularly in the Sewed-ViTraj dataset, it achieves an impressive 11.5\% improvement in the IoU metric compared to the sub-best model. This underscores the ability of data-driven models to deliver more remarkable results when contrasted with heuristic rule-based models. 
    
Among the baseline models, the HMM exhibits superior precision compared to the SP approach on both datasets. With the incorporation of the hard denoise, both methods, SP+DHM$^{\text{de}}$ and HMM+DHM$^{\text{de}}$, show a marginal enhancement in precision but a decrease in recall. {Conversely, the inclusion of tracklet information produces the reverse outcome. This suggests that employing a hard threshold for denoising unavoidably sacrifices true positive samples when identifying noise, while the tracklet leads to the model generating a more verbose trajectory when encountering noisy records.}

\begin{figure*}[t]
    \centering
    \includegraphics[width=\textwidth]{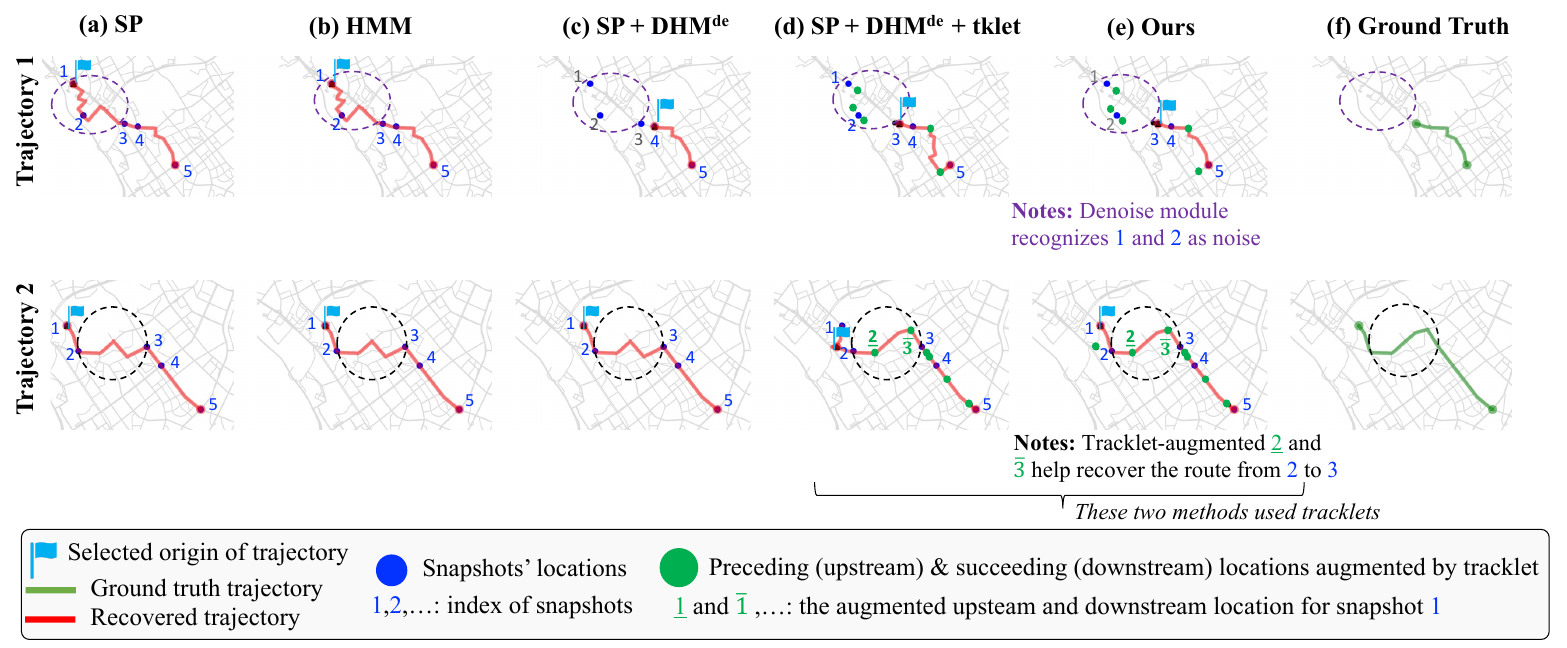}
    \caption{The case study of the recovered trajectories (red lines) conditioned on the captures (blue points) from different methods. 
    } 
    \label{fig:vis_traj}
\end{figure*}

\textbf{Case study:} 
Fig. \ref{fig:vis_traj} visualizes the reconstructed trajectories. 
Each column illustrates the outcomes derived from a specific model, and subfigures in a given row depict the results from distinct models applied to a test sample. 
The blue points indicate the locations of the cameras, and the green and red lines mean the ground truth and recovery results conditioned on these camera positions (i.e., captures), respectively.

According to Fig. \ref{fig:vis_traj} (e), our proposed method successfully reconstructs the correct trajectory despite many misleading noises and missing captures. Moreover, one can impress the superiority of our proposed two components from the presented cases:
(1) \textbf{The significance of the denoising module:}  As the violet circles show, the trajectories reconstructed by both SP and HMM (first row in Fig. \ref{fig:vis_traj} (a)(b)) deviate from the ground truth in several parts of the trajectory due to the misleading noises (false positives), while the results from the last three methods remain largely unaffected. 
{This disparity can be attributed to the noise recognition capabilities of the denoising module.} Through either soft (ours) or hard (DHM's) denoising solutions, the recovery process has successfully eschewed the false positives. 
(2) \textbf{The importance of the tracklet information:} Examining the results from SP, HMM, and SP+DHM$^\text{de}$ (second row in Fig. \ref{fig:vis_traj} (a)(b)(c)), it is evident that these methods produce incorrect detours (depicted as black zones). 
The limited coverage of deployed cameras causes the loss of surveillance information for vehicles.
From Fig. \ref{fig:vis_traj} (e), the up/downstream nodes extracted from tracklets reduce the uncertainty of the route by providing contextual knowledge and thus facilitate the recovery of a correct trajectory. 

\subsection{Ablation studies}
Given that the proposed model incorporates both a denoising module and tracklet information, we perform the ablation experiments to provide insights into the two boosting components for the objective function. The following configurations are examined:
\begin{itemize}
    \item w/o tklet: The model without the inclusion of tracklet.
    \item w/o de: The model without the denoising module, essentially a Transformer with tracklet information as input.
    \item w/o de+tklet: The model without both the denoising module and tracklet information, essentially comprising only the Transformer backbone.
\end{itemize} 

Their performance is summarized in Table \ref{tab:ablation}. The model without the denoising (w/o de) component exhibits the most significant degradation (maximum -4.0\% in IoU), underscoring the critical importance of the designed soft-denoising module. In comparison, the model without tracklet information (w/o tklet) shows inferior results (approximately -1.8\% $\sim$ -2.7\% degradation) compared to the proposed method, highlighting the beneficial impact of tracklet information on model performance. Meanwhile, the backbone model (w/o de+tklet) delivers moderate performance.
{It is worth noting that, unlike the performance decrease in certain criteria resulting from the introduction of two boosting methods for the baselines (SP and HMM) in Table {\ref{tab:overall_perf}}, our approach can circumvent this phenomenon, where the performance does not degrade with the incorporation of modules. This indicates that our proposed model can harmonize and exploit the two boosting modules.}
\begin{table}[t]
\centering
\caption{Ablation Results}
\label{tab:ablation}
\resizebox{\columnwidth}{!}{
\begin{tabular}{c|ccc|ccc}\toprule
 \multirow{2}{*}{\textbf{Metrics}}                       & \multicolumn{3}{c}{\textbf{Sewed-ViTraj}} & \multicolumn{3}{c}{\textbf{Simulated-ViTraj}} \\\cline{2-7}
                 & Precision   & Recall   & IoU    & Precision     & Recall    & IoU      \\\midrule
w/o de+tklet & 0.826       & 0.846    & 0.729    & 0.868         & 0.883     & 0.826    \\\hline
w/o tklet               & 0.836       & 0.860    & 0.749   & 0.880         & 0.904     & 0.829    \\\hline
w/o de           & 0.878       & 0.899    & 0.807    & 0.877         & 0.908     & 0.820    \\\hline
Full Model     &    0.903        &    0.917       &      0.836         &  0.896       & 0.924    & 0.852    \\\bottomrule
\end{tabular}}
\end{table}
\subsection{Practicality}
\begin{figure*}[t]
    \centering
    \includegraphics[width=1.4\columnwidth]{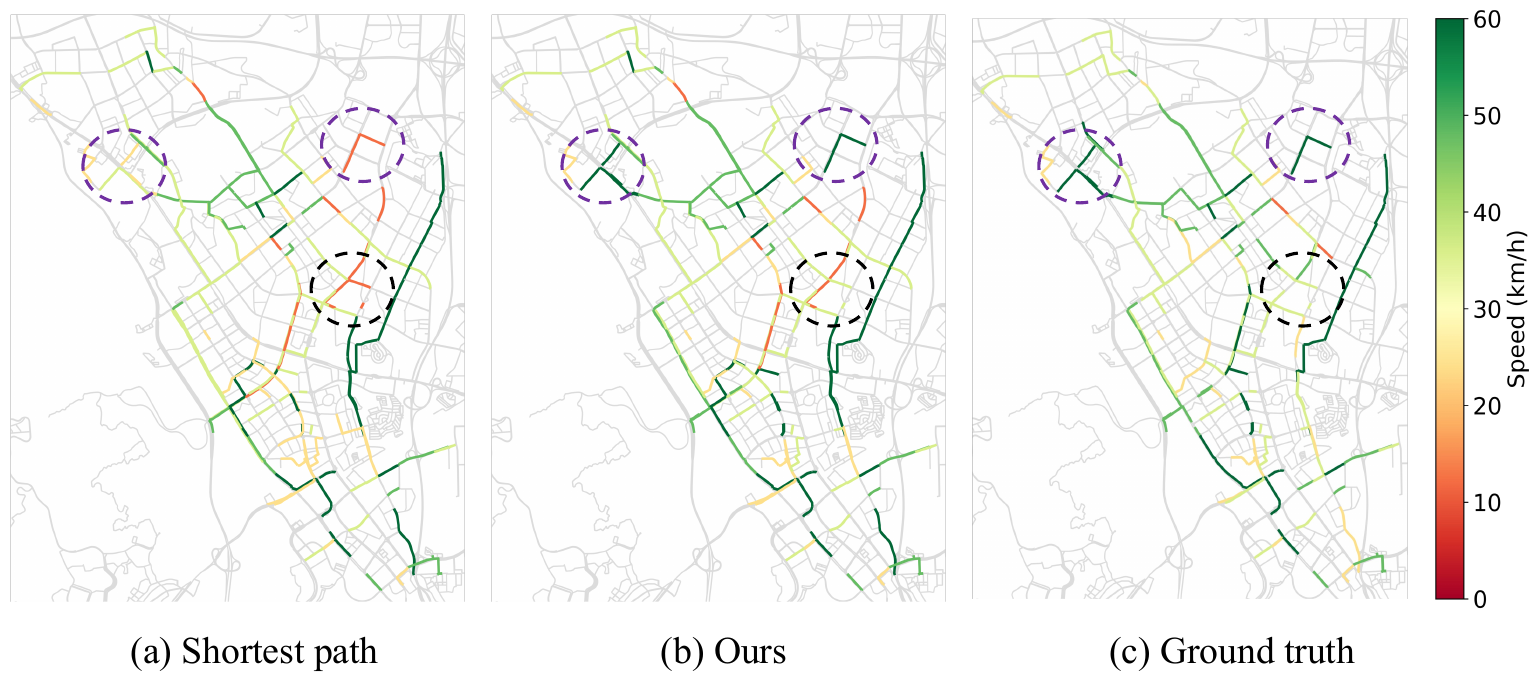}
    \caption{The road network-level travel speed calculated from trajecotries}
    \label{fig:speed}
\end{figure*}

To investigate the use cases of trajectory recovery, we finally explore and develop two trajectory-based applications, which are downstream tasks affected by recovery quality.

\begin{table}[t]
\caption{The clustering performance with and without feedback}
\centering
\begin{tabular}{cccc}\toprule
                                    & \textbf{Precision} & \textbf{Recall} & \textbf{F1-score} \\\midrule
w/o feedback                     & 0.942     & 0.483  & 0.639    \\
with feedback & 0.976     & 0.532  & 0.689   \\\bottomrule
\end{tabular}
\label{tab:fdbk}
\end{table}

\subsubsection{Multi-modal clustering performance improvement}
The recovered trajectories are used to update the clustering results, where records situated on trajectories are treated as true positives, while others are identified as noises. Through noise feedback mechanisms and the removal of noisy records, our objective is to enhance the performance of the canonical clustering approach introduced in Section \ref{sec: reid}, in which the records are grouped by examining the multi-modal similarities according to a fixed threshold (default set at 0.9).
We evaluate the canonical clustering and clustering with trajectory feedback approaches on clustering-related metrics in Table \ref{tab:fdbk}. 
Our findings indicate an overall improvement in clustering results following trajectory recovery feedback.

\subsubsection{Road network-level travel speed monitoring}

Utilizing the recorded timestamps and extracted route distances from recovered trajectories, we compute the velocities of links within the road network covered by the trajectory. For the camera-undeployed yet trajectory-covered nodes (intersections), the vehicle through-time stamps are linearly interpolated.

The calculated travel speeds are visually represented in Fig. \ref{fig:speed}. Referred to the ground truth depicted in Fig. \ref{fig:speed} (c), it reveals that the travel speeds derived from our reconstructed trajectories (Fig. \ref{fig:speed} (b)) align more closely with the facts than those obtained from the SP-derived trajectories (Fig. \ref{fig:speed} (a)). Notable improvements are highlighted within dashed circles.   In the violet zones, the speed evaluated from our reconstructed trajectories is approximately 50 km/h. In contrast, the SP outputs suggest a speed of around 20 km/h, deviating significantly from the ground truth. This discrepancy not only misrepresents the traffic conditions but also increases the risk of triggering false alarm congestion. In the black zones, although both SP outputs and ours carry the risk of producing false negatives, our method exhibits a comparatively slighter impact.

\section{Conclusion}
Vision-based trajectory recovery facilitates cost-effective travel monitoring and hot route finding in the urban city. 
This paper investigates a learning-based trajectory recovery method with vision-based snapshot input. 
Dedicated to this, we propose VisionTraj, the first-ever sequence-to-sequence framework that integrates two plug-and-play components (i.e., soft-denoising module and tracklets) into recovery. 
Coupled with it, two separate datasets are released to alleviate the scarcity of public-available related datasets, and we hope the study can foster the research community. Finally, two trajectory-based applications are also explored and developed, which demonstrate beneficial downstream use cases.

We aspire that this work motivates future research of vision-based trajectory recovery in two areas: 1) Investigating more efficient dataset construction and powerful data-driven methods for accurate recovery; and 2) Unlocking additional capabilities such as the transferability in an unseen city and the generalization of the two plug-and-play components, encouraging other researchers to explore diverse possibilities.

\section*{Aknowledgement}


Zhishuai Li and Ziyue Li led model design, project management, conducting experiments, and paper writing; Xiaoru Hu and Guoqing Du built the simulation systems of Longhua district and the vehicles; Yunhao Nie contributed to the first phase of the algorithm design; Feng Zhu and Sicheng Liu did the visual clustering pipeline; Lei Bai contributed to the ideation and paper polishing; Rui Zhao contributed to the key ideation of introducing fine- and coarse-grained denoising and tracklets.
The authors also thank the product team, which helped commercialize the model for industry applications.

\bibliographystyle{plain}
\bibliography{ref}
\end{document}